\documentclass[11pt,a4paper]{article}
\usepackage[hyperref]{acl2019}
\usepackage{times}
\usepackage{latexsym}

\usepackage{url}

\usepackage{array,multirow}
\usepackage{enumitem}
\usepackage{graphicx}
\usepackage{caption}
\usepackage{subcaption}
\usepackage{amsmath}
\usepackage{amsfonts}
\usepackage{xcolor}
\graphicspath{ {figs/} }
\newtheorem{Condition}{Condition}

\aclfinalcopy 

\title{Continual and Multi-Task Architecture Search}

\author{Ramakanth Pasunuru \and Mohit Bansal \\
  UNC Chapel Hill \\
  {\tt \{ram, mbansal\}@cs.unc.edu} \\
 }

\date{}

\begin{document}
\maketitle
\begin{abstract}
Architecture search is the process of automatically learning the neural model or cell structure that best suits the given task. Recently, this approach has shown promising performance improvements (on language modeling and image classification) with reasonable training speed, using a weight sharing strategy called Efficient Neural Architecture Search (ENAS).
In our work, we first introduce a novel continual architecture search (CAS) approach, so as to continually evolve the model parameters during the sequential training of several tasks, without losing performance on previously learned tasks (via block-sparsity and orthogonality constraints), thus enabling life-long learning. Next, we explore a multi-task architecture search (MAS) approach over ENAS for finding a unified, single cell structure that performs well across multiple tasks (via joint controller rewards), and hence allows more generalizable transfer of the cell structure knowledge to an unseen new task. We empirically show the effectiveness of our sequential continual learning and parallel multi-task learning based architecture search approaches on diverse sentence-pair classification tasks (GLUE) and multimodal-generation based video captioning tasks. Further, we present several ablations and analyses on the learned cell structures.\footnote{All our code and models publicly available at: \url{https://github.com/ramakanth-pasunuru/CAS-MAS}}

\end{abstract}

\section{Introduction}

\label{sec:introduction}
Architecture search enables automatic ways of finding the best model architecture and cell structures for the given task or dataset, as opposed to the traditional approach of manually choosing or tuning among different architecture choices, which introduces human inductive bias or is non-scalable. Recently, this idea has been successfully applied to the tasks of language modeling and image classification~\cite{zoph2016neural,zoph2017learning,cai2018efficient,liu2017progressive,liu2017hierarchical}. The first approach of architecture search involved an RNN controller which samples a model architecture and uses the validation performance of this architecture trained on the given dataset as feedback (or reward) to sample the next architecture. 
Some recent attempts have made architecture search more computationally feasible~\cite{negrinho2017deeparchitect,baker2016designing} via tree-structured search space or Q-learning with an $\epsilon$-greedy exploration, and further improvements via a  weight-sharing strategy called Efficient Neural Architecture Search (ENAS)~\cite{pham2018efficient}.

In this work, we extend the architecture search approach to an important paradigm of transfer learning across multiple data sources: \emph{continual learning}.
The major problem in continual learning is \emph{catastrophic forgetting}. 
For this, we introduce a novel `continual architecture search' (CAS) approach, where the model parameters evolves and adapts when trained sequentially on a new task while maintaining the performance on the previously learned tasks. For enabling such continual learning, we formulate a two-step graph-initialization approach with conditions based on block sparsity and orthogonality. 
Another scenario of transfer learning or generalization that we explore is one in which we are given multiple tasks in parallel and have to learn a single cell that is good at all these tasks, and hence allows more generalizable transfer of the cell structure knowledge to a new unseen task. This is inspired by the traditional LSTM cell's reasonable performance across a wide variety of tasks, and hence we want to automatically search (learn) a better version of such a generalizable single cell structure, via multi-task architecture search (MAS). We achieve this by giving a joint reward from multiple tasks as feedback to the controller. Hence, overall, we present two generalization approaches: CAS learns generalizable model parameters over sequential training of multiple tasks (continual learning), whereas MAS learns a generalizable cell structure which performs well across multiple tasks.

For empirical evaluation of our two approaches of continual and multi-task cell learning, we choose three domains of natural language inference (NLI) bi-text classification tasks from the GLUE benchmark~\cite{wang2018glue}: QNLI, RTE, and WNLI, and three domains of multimodal-generation based video captioning tasks: MSR-VTT~\cite{xu2016msr}, MSVD~\cite{chen2011collecting}, and DiDeMo~\cite{hendricks2017localizing}. Note that we are the first ones to use the architecture search approach for text classification tasks as well as multimodal conditioned-generation tasks, which achieves improvements on the strong GLUE and video captioning baselines. 

Next, for continual learning, we train the three tasks sequentially for both text classification and video captioning (through our continual architecture search method) and show that this approach tightly maintains the performance on the previously-learned domain (also verified via human evaluation), while also significantly maximizing the performance on the current domain, thus enabling life-long learning~\cite{chen2016lifelong}. For multi-task cell learning, we show that the cell structure learned by jointly training on the QNLI and WNLI tasks, performs significantly better on the RTE dataset than the individually-learned cell structures. Similarly, we show that the cell structure learned from jointly training on the MSR-VTT and MSVD video captioning datasets performs better on the DiDeMo dataset than the individually-learned cell structures. 
Finally, we also present various analyses for the evolution of the learned cell structure in the continual learning approach, which preserves the properties of certain edges while creating new edges for new capabilities. For our multi-task learning approach, we observe that the joint-reward cell is relatively less complex than the individual-task cells in terms of the number of activation functions, which intuitively relates to better generalizability.


\section{Related Work}
\label{sec:related-work}
Neural architecture search (NAS) has been recently introduced for automatic learning of the model structure for the given dataset/task~\cite{zoph2016neural,zoph2017learning}, and has shown good improvements on image classification and language modeling. NAS shares some similarity to program synthesis and inductive programming~\cite{summers1986methodology,biermann1978inference}, and it has been successfully applied to some simple Q\&A tasks~\cite{liang2010learning,neelakantan2015neural,andreas2016learning,lake2015human}.
NAS was made more computationally feasible via tree-structured search space or Q-learning with $\epsilon$-greedy exploration strategy and experience replay~\cite{negrinho2017deeparchitect,baker2016designing}, or a weight-sharing strategy among search space parameters called Efficient Neural Architecture Search (ENAS)~\cite{pham2018efficient}. 
We explore architecture search for text classification and video caption generation tasks and their integration to two transfer learning paradigms of continual learning and multi-task learning.

The major problem in continual learning is catastrophic forgetting. Some approaches addressed this by adding regularization to penalize functional or shared parameters' change and learning rates~\cite{razavian2014cnn,li2017learning,hinton2015distilling,jung2016less,kirkpatrick2017overcoming,donahue2014decaf,yosinski2014transferable}.
Others proposed copying the previous task and augmenting with new task's features~\cite{rusu2016progressive}, intelligent synapses to accumulate task-related information~\cite{zenke2017continual}, or online variational inference~\cite{nguyen2017variational}. Also,~\newcite{yoon2018lifelong} proposed a dynamically expandable network based on incoming new data. In our work, we introduce `continual architecture search' by extending the NAS paradigm to avoid catastrophic forgetting via block-sparsity and orthogonality constraints, hence enabling a form of life-long learning~\cite{chen2016lifelong}. To the best of our knowledge, our paper is the first to extend architecture search to a continual incoming-data setup.~\newcite{elsken2018efficient} and~\newcite{so2019evolved} proposed evolutionary architecture search algorithms that dynamically allocate more resources for promising architecture candidates, but these works are different from us in that they do not consider the case where we have continual incoming-data from different data sources, but instead focus on the continual evolution of the model search for efficiency purposes.

Multi-task learning (MTL) is primarily used to improve the generalization performance of a task by leveraging knowledge from related tasks~\cite{caruana1998multitask,collobert2008unified,girshick2015fast,luong2015multi,ruder2017sluice,augenstein2018multi,guo2018soft,oh2017zero,ruder2017learning}.
In similar generalization spirit of multi-task learning, we present multi-task architecture learning based on performance rewards from multiple tasks, so as to find a  single cell structure which can generalize well to a new unseen task.


\section{Architecture Search for Text Classification and Generation}
In this section, we first discuss how we adapt ENAS~\cite{pham2018efficient} for modeling our bi-text classification and multimodal video captioning tasks. Next, we introduce our continual and multi-task approaches of transfer learning leveraging architecture search.

\subsection{ENAS Algorithm}
\label{subsec:enas}
\begin{figure}[t]
\centering
\begin{subfigure}{.85\linewidth}
  \centering
  \includegraphics[width=0.85\linewidth]{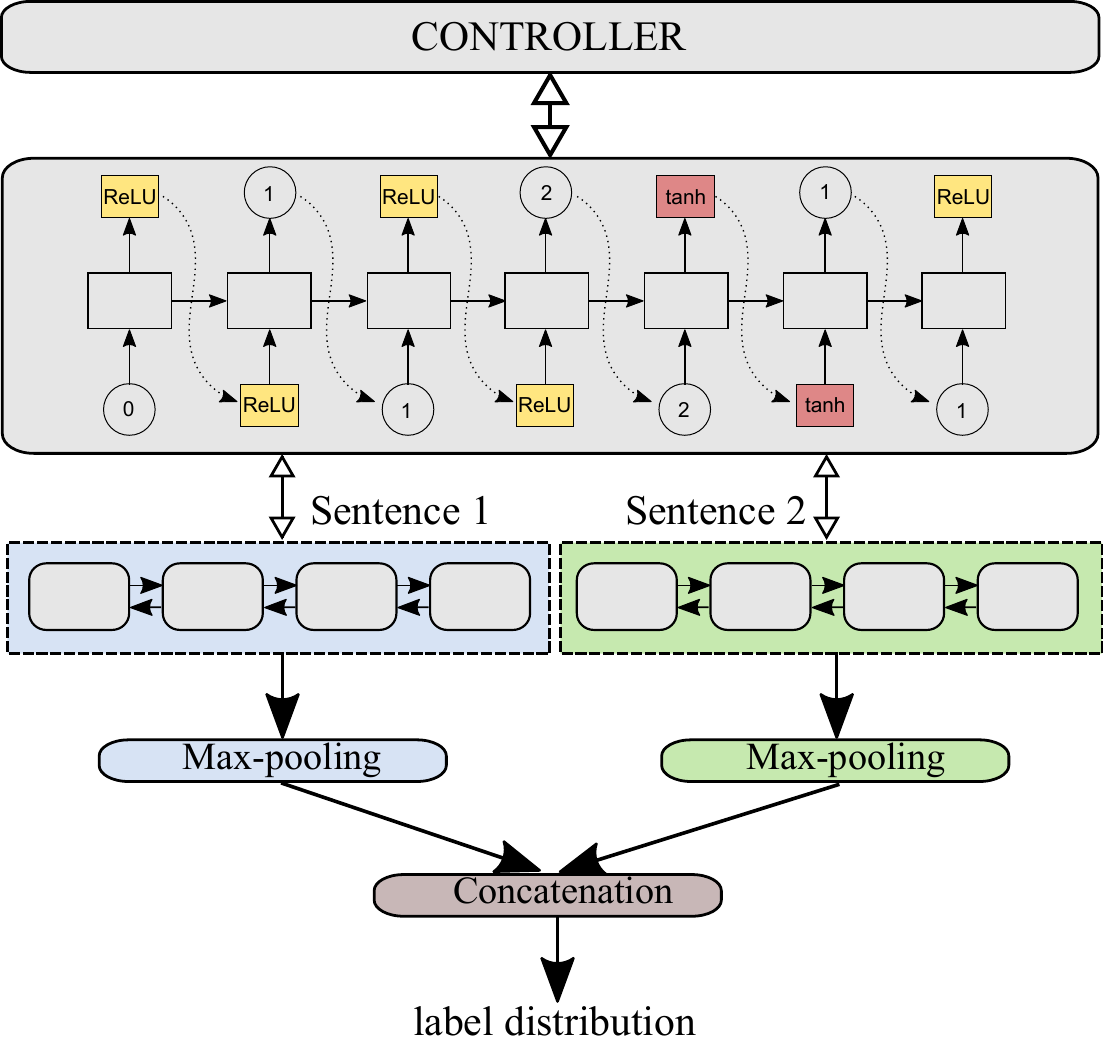}
  \caption{Text classification ENAS.\label{fig:enas-classifcation}}
\end{subfigure}
\\
\begin{subfigure}{.85\linewidth}
  \centering
  \vspace{10pt}
  \includegraphics[width=0.85\linewidth]{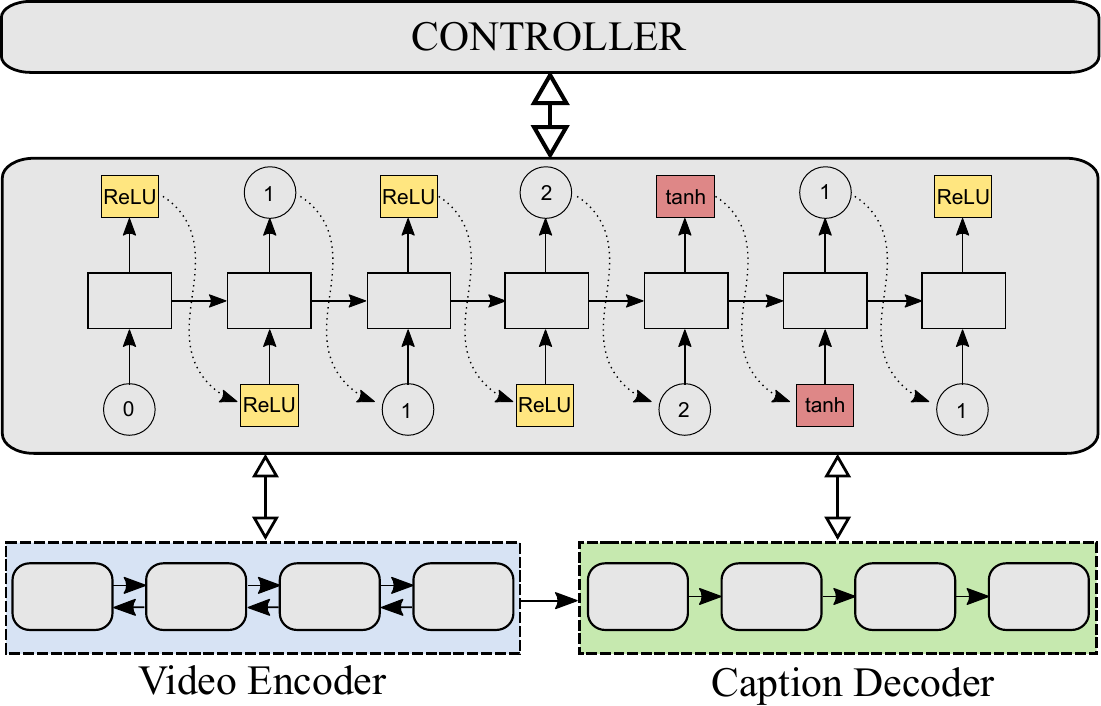}
  \caption{Video captioning ENAS.\label{fig:enas-generation}}
\end{subfigure}
\vspace{-7pt}
\caption{Architecture search models for bi-text classification and video caption generation tasks.
\label{fig:enas-models}
\vspace{-7pt}
}
\end{figure}

Our initial architecture search approach is based on the recent Efficient Neural Architecture Search (ENAS) method of~\newcite{pham2018efficient}, but modeled for text classification and generation-based video captioning. Fig.~\ref{fig:enas-models} presents the ENAS controller for sampling an RNN cell structure, which we use to learn the two encoders of our text classification model or encoder-decoder for our video captioning model. The controller is a simple LSTM-RNN and the classifier encoder's or video captioning encoder-decoder's RNN cell structure is based on the combination of $N$ nodes indexed by $h_1^{(t)}, h_2^{(t)},.., h_N^{(t)}$ (edges between nodes represent weight parameters) and activation functions (ReLU, tanh, sigmoid, identity), where $t$ denotes the time step.
For node $h_1^{(t)}$, there are two inputs: $x^{(t)}$ (input signal) and $h_N^{(t-1)}$ (output from previous time-step), and the node computations are:
\vspace{-2pt}
\begin{equation}
c_1^{(t)} = \mathrm{sigmoid}(x^{(t)} \cdot W^{(x,c)} + h_N^{(t-1)} \cdot W_0^{(c)})
\vspace{-7pt}
\end{equation}
\vspace{-14pt}
\begin{equation}
\begin{split}
h_1^{(t)} &= c_1^{(t)}{\odot}f_1(x^{(t)}{\cdot}W^{(x,h)}{+}h_N^{(t-1)}{\cdot}W_1^{(h)}) \\
 & + (1-c_1^{(t)}) \odot h_N^{(t-1)}
\end{split}
\end{equation}
where $f_1$ is the activation function. Node $h_l$, where $l \in \{2,3,..,N\}$, receives input from node $j_l$ where $j_l \in \{h_1, h_2,..,h_{l-1}\}$, and the computation is defined as follows:
\begin{equation}
c_l^{(t)} = \mathrm{sigmoid}(h_{j_l}^{(t)} \cdot W^{(c)}_{l,j_l} )
\vspace{-10pt}
\end{equation}
\begin{equation}
h_l^{(t)} = c_l^{(t)} \odot f_l(h^{(t)}_{j_l} \cdot W^{(h)}_{l,j_l})
+ (1-c_l^{(t)}) \odot h_{j_l}^{(t)}
\end{equation}
During training, we alternately train the model parameters and controller parameters. First, we sample a Directed Acyclic Graph (DAG) structure from the controller at every mini-batch and use it to update the weight parameters of the task's RNN nodes/parameters. Next, we sample a DAG from the controller and measure the (validation) performance of that structure based on this new updated state of the task model, and use this performance as a reward to allow the controller to update its own parameters. We repeat this alternate training procedure until the model converges. Later, we select the DAG structure with the best performance and use it to retrain the model from scratch.

\subsection{ENAS for Bi-Text Classification}
\label{subsec:text-classification}
For our NLI text classification tasks, we are given the sentence pair as input, and we have to classify it as entailment or not. For a strong base model, we follow~\newcite{conneau2017supervised} model, and use bidirectional LSTM-RNN encoders to encode both the sentences and then we do max-pooling on the outputs from these encoders. Let $v$ represent the max-pooling output from the first sentence encoder and $u$ represent the max-pooling output from the second sentence encoding. The joint representation $h$ is defined as $h = [u ; v; |u-v|; u \odot v]$. The final representation is linearly projected to the label classes, and then fed through softmax to get the final class distribution.
Fig.~\ref{fig:enas-classifcation} presents an overview of our text classification model along with ENAS controller for sampling an RNN cell structure. We sample an RNN cell structure from the ENAS controller and use it in the two recurrent encoders of the bi-text classification model. In the first stage, we learn the best cell structure, by sampling multiple cell structures and giving the corresponding validation accuracy as the feedback reward to the controller. In the second stage, we use the best cell structure from the stage-1 to retrain the text classification model from scratch.

\subsection{ENAS for Conditioned Generation}
\label{subsec:seq2seq-model}

Next, we go beyond text classification, and look at conditioned text generation with ENAS, where we choose the task of video-conditioned text generation (also known as video captioning) so as to also bring in a multi-modality aspect.
For a strong baseline, we use a sequence-to-sequence model with an attention mechanism similar to~\newcite{pasunuru2017multitask}, where we encode the video frames as a sequence into a bidirectional LSTM-RNN and decode the caption through another LSTM-RNN (see Fig.~\ref{fig:enas-generation}). Our attention mechanism is similar to~\newcite{bahdanau2014neural}, where at each time step $t$ of the decoder, the LSTM hidden state $s_t$ is a non-linear function of previous time step's decoder hidden state $s_{t-1}$ and generated word $w_{t-1}$, and the context vector $c_t$ which is a weighted combination of the encoder hidden states $\{h^i\}$. These weights $\alpha_{t}$, are defined as:

\begin{equation}
    \alpha_{t,i} = \frac{\exp(e_{t,i})}{\sum_{k=1}^n \exp(e_{t,k})}
\end{equation}

The attention function $e_{t,i} = w^{T} \tanh(W_ah_i + U_as_{t-1}+b_a)$, where $w$, $W_a$, $U_a$, $b_a$ are learned parameters. Fig.~\ref{fig:enas-generation} presents our video captioning model along with ENAS controller.
Here, we sample an RNN cell structure from the ENAS controller and use it for both encoder and decoder, and rest of the ENAS procedure is similar to Sec.~\ref{subsec:text-classification}.

\section{Continual Architecture Search (CAS)}
\label{subsec:continual-learning}

\begin{figure*}
\centering
\includegraphics[width=0.87\linewidth]{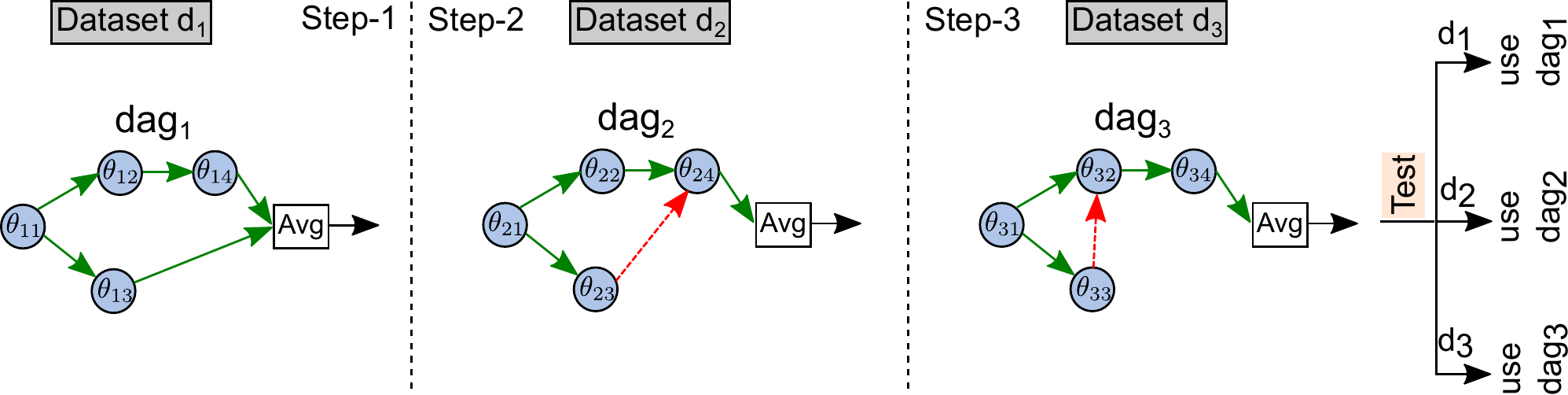}
\vspace{-10pt}
\caption{Continual architecture search (CAS) approach: green, solid edges (weight parameters) are shared, newly-learned edges are represented with red, dashed edges.
}
\label{fig:alcl}
\vspace{-10pt}
\end{figure*}

We introduce a novel continual learning paradigm on top of architecture search, where the RNN cell structure evolves when trained on new incoming data/domains, while maintaining the performance on previously learned data/domains (via our block-sparsity and orthogonality conditions discussed below), thus enabling life-long learning~\cite{chen2016lifelong}.
Let $\theta_{1,k} \in \theta_1$ and $\theta_{2,k} \in \theta_2$ (where $k$ denotes model parameters) be the learned model parameters for task $T$ when independently trained on datasets $d_1$ and $d_2$. Then, we can say that $\theta_{2,k} = \theta_{1,k} +  \psi_{2,k}$, 
where, $\psi_{2,k}$ is the change in the model parameters of $\theta_{1,k}$ when trained independently on $d_2$. There are infinitely many possible local optimal solutions for $\psi_{2,k}$, hence in our continual learning approach, we want to learn the parameters $\psi_{2,k}$ when training on dataset $d_2$ such that it will not affect the performance of the task w.r.t. dataset $d_1$. For this, we formulate two important conditions:
\begin{Condition} When training the model on dataset $d_1$, we constrain the model parameters $\theta_{1,k} \in \mathrm{R}^{m \times n}$ to be sparse, specifically, to be block sparse, i.e., minimize $ \sum_{i=1}^m | (||\theta_{1,k}[i,:]||_2) |_1$. 
\end{Condition}
Here, $||\cdot||_2$ represents the $l_2$ norm and $||\cdot||_1$ represents the $l_1$ norm. $l_2$ and $l_1$ norms are efficient in avoiding over-fitting; however, they are not useful for compact representation of the network.~\newcite{scardapane2017group} proposed group sparsity in the neural networks to completely disconnect some neurons. Our block sparse condition is inspired from their work. 
This sparsity condition is also useful for our continual learning approach which we discuss in Condition 2.
\begin{Condition}
When training the model on dataset $d_2$, we start from $\theta_{1,k}$, keep it constant, and update $\psi_{2,k}$ such that:
\begin{enumerate}[topsep=0pt,itemsep=-1ex,partopsep=1ex,parsep=1ex]
\item $\psi_{2,k}$ is block sparse, i.e., minimize $ \sum_{i=1}^m | (||\psi_{2,k}[i,:]||_2) |_1$.
\item $\theta_{1,k}$ and $\psi_{2,k}$ are orthogonal.
\end{enumerate}
\end{Condition}

It is important in the continual learning paradigm that we do not affect the previously learned knowledge. As stated in Condition 1, we find a block sparse solution $\theta_{1,k}$ such that we find the solution $\theta_{2,k}$ which is close to $\theta_{1,k}$ and the new knowledge is projected in orthogonal direction via $\psi_{2,k}$ so that it will not affect the previously learned knowledge, and thus `maintain' the performance on previously learned datasets. 
We constrain the closeness of $\theta_{2,k}$ and $\theta_{1,k}$ by constraining $\psi_{2,k}$ to also be block sparse (Condition 2.1). 
Also, to avoid affecting previously learned knowledge, we constrain $\theta_{1,k}$ and $\psi_{2,k}$ to be orthogonal (Condition 2.2). However, strictly imposing this condition into the objective function is not feasible~\cite{bousmalis2016domain}, hence we add a penalizing term into the objective function as an approximation to the orthogonality condition: $L_p(\theta_{2,k}) = ||\theta_{1,k}^T \cdot \psi_{2,k} ||^2_2 $. Both Condition 2.1 and 2.2 are mutually dependent, because for two matrices' product to be zero, they share basis vectors between them, i.e., for an $n$-dimensional space, there are $n$ basis vectors and if $p$ of those vectors are assigned to one matrix, then the rest of the $n-p$ vectors (or subset) should be assigned to the other matrix.\footnote{Note that it is not necessary for the matrix to contain all of the $n-p$ basis vectors, if the matrix rank is less than $n$, then it may have less than $n-p$ basis vectors.} If we fill the rest of the rows with zeros, then they are block sparse, which is the reason for using Condition 2.1. Our CAS condition ablation (see Sec.~\ref{subsec:continual-learning-on-glue}) shows that both these conditions are necessary for continual learning.

Next, we describe the integration of our above continual learning approach with architecture search, where the model continually evolves its cell architecture so as to perform well on the new incoming data, while also tightly maintaining the performance on previously learned data (or domains). 
Fig.~\ref{fig:alcl} presents an overview of our continual learning integration approach into architecture search for sequential training on three datasets. 
Initially, given the dataset $d_1$, we train the architecture search model to find the best Directed Acyclic Graph (DAG) structure for RNN cell and model parameters $\theta_{1,k}$ under the block sparse condition described above in Sec.~\ref{subsec:continual-learning}. We call this step-1, corresponding to dataset $d_1$. Next, when we have a new dataset $d_2$ from a different domain, we further continue to find the best DAG and model parameters $\theta_{2,k}$ for best performance on $d_2$, but initialized the parameters with step-1's parameters $\theta_{1,k}$, and then trained on dataset $d_2$ following Condition 2 (discussed in Sec.~\ref{subsec:continual-learning}). We call this step-2, corresponding to dataset $d_2$. After the end of step-2 training procedure, for re-evaluating the model's performance back on dataset $d_1$, we still use the final learned model parameters $\theta_{2,k}$, but with the learned DAG from step-1.\footnote{For evaluating the model's performance on dataset $d_2$, we obviously use the final learned model parameters $\theta_{2,k}$, and the learned DAG from step-2.} This is because we cannot use the old step-1 model parameters $\theta_{1,k}$ since we assume that those model parameters are not accessible now (assumption for continual learning with large incoming data streams and memory limit for saving large parameter sets). 

\begin{figure}
\centering
\includegraphics[width=0.7\linewidth]{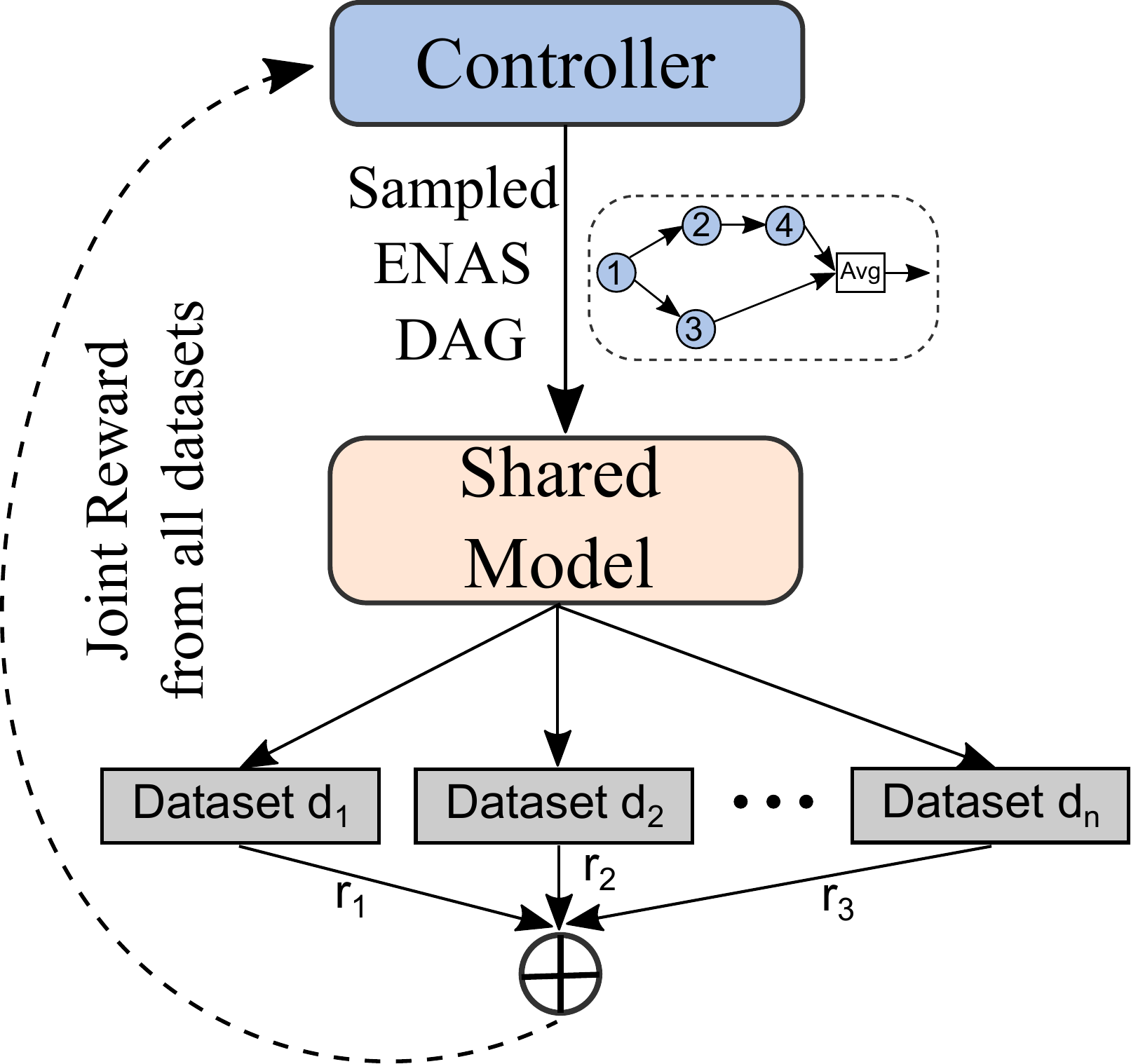}
\vspace{-10pt}
\caption{Multi-task cell structure learning using joint rewards from $n$ datasets.
}
\label{fig:multitask-cell-model}
\vspace{-10pt}
\end{figure}

\section{Multi-Task Architecture Search (MAS)}
\label{sec:multitask-enas}
In some situations of transfer learning, we are given multiple tasks at once instead of sequentially. In such a scenario, when we train architecture search model on these multiple tasks separately, we get different cell structures on each task which overfit to that task and are not well generalizable. So, instead, we should learn a common cell for multiple tasks which should generalize better to an unseen task.
Also, the standard non-architecture search based LSTM-RNN cell performs well across different tasks which shows enough evidence that there exist such architectures that work well across different tasks. Hence, in our work, we aim to follow a data-driven route to find even better generalizable architectures that perform better than the traditional LSTM-RNN cell, via our multi-task architecture search (MAS) approach, described below. 

To learn a cell architecture on a task, we provide the performance of the sampled cell structure on the validation set of the given task as reward to the controller. However, our aim is to find a generalizable cell structure which jointly performs well across different tasks/datasets $\{d_1, d_2,.., d_n\}$. Hence, during the architecture search training, the joint reward to the controller is a combination of the performance scores of the sampled cell structure on the validation set of all the available/candidate tasks, which is defined as $ r_c = \frac{1}{n} \sum_{i=1}^n r_i $, where reward $r_i$ comes from the validation performance on task/dataset $d_i$. Next, for fair generalizability comparison of this multi-task cell structure with other individual task-learned cell structures, we choose a new unseen task which is different from the current candidate tasks and show that the multi-task cell performs better on this unseen task than all task-related cell structures (as well as a non-ENAS LSTM cell).

\section{Experimental Setup}
\subsection{Text Classification Datasets}

We choose the natural inference datasets of QNLI, RTE, and WNLI from the  GLUE~\citep{wang2018glue} benchmark to perform experiments for multi-task cell structure and continual architecture search. We use the standard splits provided by~\cite{wang2018glue}.

\noindent\textbf{QNLI Dataset:} Question-Answering Natural Language Inference (QNLI) is extracted from the Stanford Question Answering Dataset~\cite{rajpurkar2016squad}, where they created sentence pair classification task by forming a pair between each question and the corresponding sentence containing the answer. Hence the task is to find whether the given sentence context contains the answer for the given question. In this dataset, we use the standard splits, i.e., 108k examples for training, 5.7k for validation, and 5.7k for testing.

\noindent\textbf{RTE Dataset:} Recognizing Textual Entailment (RTE) is collected from a series of annual challenges on the task of textual entailment. This dataset spans the news and Wikipedia text. Here, the task is to predict whether the sentence pair is entailment or not. In this dataset, we use the standard splits, i.e., 2.5k examples for training, 276 for validation, and 3k for testing.

\noindent\textbf{WNLI Dataset:}
Winograd Natural Language Inference (WNLI) is extracted from the dataset of Winograd Schema Challenge for reading comprehension task. Original dataset is converted into a sentence pair classification task by replacing the ambiguous pronoun with each possible referent, where the task is to predict if the sentence with the substituted pronoun is entailed by the original sentence. We use 634 examples for training, 71 for validation, and 146 for testing.

\subsection{Video Captioning Datasets}
For the conditioned-generation paradigm, we use three popular multimodal video captioning datasets: MSR-VTT, MSVD, and DiDeMo to perform experiments for continual architecture search and multi-task architecture search. 

\noindent\textbf{MSR-VTT Dataset:}
MSR-VTT is a collection of $10,000$ short videos clips collected from a commercial search engine covering $41.2$ hours of video and annotated through Amazon Mechanical Turk (AMT). Each video clip has $20$ human annotated captions. We used the standard splits following previous work, i.e., $6,513$ video clips as training set, $497$ as validation set, and $2,990$ as test set.

\noindent\textbf{MSVD Dataset:}
Microsoft Video Description Corpus (MSVD) is a collection of $1970$ short video clips collected in the wild and annotated through Amazon Mechanical Turk (AMT) in different languages. In this work, we use only English language annotations. Each video clip on an average is $10$ seconds in length and approximately $40$ annotations. We use the standard splits following previous work, i.e., $1,200$ video clips as training set, $100$ as validation set, and $670$ as test set. 

\noindent\textbf{DiDeMo Dataset:}
Distinct Describable Moments (DiDeMo) is traditionally a video localization task w.r.t. given description query~\cite{hendricks2017localizing}. In this work, we use it as a video description task where given the video as input we have to generate the caption. We use the standard splits as provided by~\newcite{hendricks2017localizing}.

\subsection{Evaluation}
For GLUE tasks, we use accuracy as an evaluation metric following the previous work~\cite{wang2018glue}. For video captioning tasks, we report four diverse automatic evaluation metrics: METEOR~\cite{banerjee2005meteor}, CIDEr~\cite{vedantam2015cider}, BLEU-4~\cite{papineni2002bleu}, and ROUGE-L~\cite{lin2004rouge}. We use the standard evaluation code~\cite{chen2015microsoft} to obtain these scores for our generated captions w.r.t. the reference captions.

\subsection{Training Details}
In all our experiments, our hyperparameter choices are based on validation set accuracy for GLUE tasks and an average of the four automatic evaluation metrics (METEOR, CIDEr, BLEU-4, and ROUGE-L) for video captioning tasks. We use same settings for both normal and architecture search models, unless otherwise specified. More details in appendix.


\section{Results and Analysis}
\label{sec:results}

\subsection{Continual Learning on GLUE Tasks}
\label{subsec:continual-learning-on-glue}

\noindent\textbf{Baseline Models:}
We use bidirectional LSTM-RNN encoders with max-pooling~\cite{conneau2017supervised} as our baseline.\footnote{We also tried various other models e.g., self-attention and cross-attention, but we found that the max-pooling approach performed best on these datasets.} Further, we used the ELMo embeddings~\cite{peters2018deep} as input to the encoders, where we allowed to train the weights on each layer of ELMo to get a final representation. Table~\ref{table:continual-learning-glue-results} shows that our baseline models achieve strong results when compared with GLUE benchmark baselines~\cite{wang2018glue}.\footnote{We only report single-task (and not 9-task multi-task) results from the GLUE benchmark for fair comparison to our models (even for our multi-task-cell learning experiments in Sec.~\ref{subsec:multi-task-cell-glue}, the controller uses rewards from two datasets but the primary task is then trained only on its own data).} On top of these strong baselines, we add ENAS approach.

\noindent\textbf{ENAS Models:} 
Next, Table~\ref{table:continual-learning-glue-results} shows that our ENAS models (for all three tasks QNLI, RTE, WNLI) perform better or equal than the non-architecture search based models.\footnote{On validation set, our QNLI ENAS model is statistically significantly better than the corresponding baseline with $p<0.01$, and statistically equal on RTE and WNLI (where the validations sets are very small), based on the bootstrap test~\cite{noreen1989computer,efron1994introduction} with 100K samples. Since the test set is hidden, we are not able to calculate the statistical significance on it.} Note that we only replace the LSTM-RNN cell with our ENAS cell, rest of the model architecture in ENAS model is same as our baseline model.\footnote{Note that ENAS random search baseline vs. optimal search validation performance on QNLI, RTE, and WNLI are 73.3 (vs. 74.8), 58.8 (vs. 60.3), and 54.0 (vs. 55.6), respectively, suggesting that the learned optimal cell structure is better than the random cell structure.}  

\noindent\textbf{CAS Models:}
Next, we apply our continual architecture search (CAS) approach on QNLI, RTE, and WNLI, where we sequentially allow the model to learn QNLI, RTE, and WNLI (in the order of decreasing dataset size, following standard transfer setup practice) and the results are as shown in Table~\ref{table:continual-learning-glue-results}. 
We train on QNLI task, RTE task, and WNLI task in step-1, step-2, and step-3, respectively. We observe that even though we learn the models sequentially, we are able to maintain performance on the previously-learned QNLI task in step-2 (74.1 vs. 74.2 on validation set which is statistically equal, and 73.6 vs. 73.8 on test).\footnote{Note that there is a small drop in QNLI performance for CAS Step-1 vs. ENAS (74.5 vs. 73.8); however, this is not true across all experiments, e.g., in case of RTE, CAS Step-1 is in fact better than its corresponding ENAS model (ENAS: 52.9 vs. CAS Step-1: 53.8).} Note that if we remove our sparsity and orthogonality conditions (Sec.~\ref{subsec:continual-learning}), the step-2 QNLI performance drops from 74.1 to 69.1 on validation set, demonstrating the importance of our conditions for CAS (see next paragraph on `CAS Condition Ablation' for more details). Next, we observe a similar pattern when we extend CAS to the WNLI dataset (see step-3 in Table~\ref{table:continual-learning-glue-results}), i.e, we are still able to maintain the performance on QNLI (as well as RTE now) from step-2 to step-3 (scores are statistically equal on validation set).\footnote{On validation set, QNLI step-3 vs. step-2 performance is 73.9 vs. 74.1, which is stat. equal. Similarly, on RTE, step-3 vs. step-2 performance is 61.0 vs. 60.6 on validation set, which is again statistically equal.} Further, if we compare the performance of QNLI from step-1 to step-3, we see that they are also stat. equal on val set (73.9 vs. 74.2). This shows that our CAS method can maintain the performance of a task in a continual learning setting with several steps. 

\begin{table}
\small
\begin{center}
\begin{tabular}{|l|c|c|c|}
\hline
Models & QNLI & RTE & WNLI \\
\hline
\multicolumn{4}{|c|}{\textsc{Previous Work}} \\
\hline
BiLSTM+ELMo~\shortcite{wang2018glue} & 69.4 &  50.1 & 65.1\\ 
BiLSTM+ELMo+Attn~\shortcite{wang2018glue} & 61.1 &  50.3 & 65.1\\ 
\hline
\multicolumn{4}{|c|}{\textsc{Baselines}} \\
\hline
Baseline (with ELMo) & 73.2 &  52.3 & 65.1 \\ 
ENAS (Architecture Search) & 74.5 & 52.9 & 65.1 \\ 
\hline
\multicolumn{4}{|c|}{\textsc{CAS Results}} \\
\hline
CAS Step-1 (QNLI training)& 73.8 & N/A  & N/A\\ 
CAS Step-2 (RTE training) & 73.6 & 54.1 & N/A \\ 
CAS Step-3 (WNLI training) & 73.3 & 54.0 & 64.4 \\ 
\hline
\end{tabular}
\end{center}
\vspace{-10pt}
\caption{Test results on GLUE tasks for various models: Baseline, ENAS, and CAS (continual architecture search). The CAS results maintain statistical equality across each step.
\label{table:continual-learning-glue-results}
\vspace{-10pt}
}
\end{table}

\noindent\textbf{CAS Condition Ablation:}
We also performed important ablation experiments to understand the importance of our block sparsity and orthogonality conditions in the CAS approach (as discussed in Sec.~\ref{subsec:continual-learning}). Table~\ref{table:cas-ablation} presents the ablation results of QNLI in step-2 with CAS conditions. Our full model (with both Condition 2.1 and 2.2) achieves a validation performance of 74.1. Next, we separately experimented with each of Condition 2.1 and 2.2 and observe that using only one condition at a time is not able to maintain the performance w.r.t. step-1 QNLI performance (the decrease in score is statistically significant), suggesting that both of these two conditions are important for our CAS approach to work. Further, we remove both conditions and observe that the performance drops to 69.1. Finally, we also replaced the QNLI cell structure with the RTE cell structure along with removing both conditions and the performance further drops to 54.1. This shows that using the cell structure of the actual task is important.

\noindent\textbf{Time Comparison:}
We compare QNLI training time on a 12GB TITAN-X Nvidia GPU. Our baseline non-ENAS model takes 1.5 hours, while our CAS (and MAS) models take approximately the same training time (4 hours) as the original ENAS setup, and do not add extra time complexity.

\begin{table}[t]
\small
\begin{center}
\begin{tabular}{|l|c|}
\hline
Model &  Accuracy on QNLI \\
\hline
No Condition with RTE DAG & 54.1 \\
No Condition & 69.1 \\
Only Condition 2.1 & 71.5 \\ 
Only Condition 2.2 & 69.4  \\ 
\hline
Full Model (Condition 2.1 \& 2.2) & 74.1 \\
\hline
\end{tabular}
\end{center}
\vspace{-10pt}
\caption{Ablation (val) results on CAS conditions. 
\label{table:cas-ablation}
}
\vspace{-17pt}
\end{table}

\begin{table*}
\begin{center}
\begin{small}
\begin{tabular}{|l|ccccc|ccccc|}
\hline
\multirow{2}{*}{Models} & \multicolumn{5}{c}{MSR-VTT} & \multicolumn{5}{|c|}{MSVD} \\
\cline{2-11}
 & C & B & R  & M & AVG & C & B & R & M & AVG\\
\hline
Baseline~\cite{pasunuru2017reinforced}  &  48.2 & 40.8 & 60.7 & 28.1 & 44.5 & 85.8 & 52.5 & 71.2 & 35.0 & 61.1 \\ 
ENAS  &  48.9 & 41.3 & 61.2  & 28.1 & 44.9 & 87.2 & 52.9 & 71.7 &  35.2 & 61.8  \\ 
\hline
CAS Step-1 (MSR-VTT training) &  48.9 & 41.1 & 60.5 &  27.5 & 44.5 & N/A & N/A & N/A & N/A & N/A \\ 
CAS Step-2 (MSVD training) &  48.4 & 40.1 & 59.9 & 27.1 & 43.9  & 88.1 & 52.4 & 71.3 & 35.1 & 61.7 \\ 
\hline
\end{tabular}
\end{small}
\end{center}
\vspace{-10pt}
\caption{Video captioning results with Baseline, ENAS, and CAS models. Baseline is reproduced numbers from github of~\newcite{pasunuru2017reinforced} which uses advanced latest visual features (ResNet-152 and ResNeXt-101) for video encoder. C, B, R, M: CIDEr, BLEU-4, ROUGE-L, and METEOR metrics.
\label{table:continual-learning-videocaptioning-results}
\vspace{-10pt}
}
\end{table*}

\subsection{Continual Learning on Video Captioning}
\label{subsec:continual-learning-on-video-captioning}
\noindent\textbf{Baselines Models:}
Our baseline is a sequence-to-sequence model with attention mechanism as described in Sec.~\ref{subsec:seq2seq-model}. We achieve comparable results w.r.t. SotA (see Table~\ref{table:continual-learning-videocaptioning-results}), hence serving as a good starting point for the ENAS approach.

\noindent\textbf{ENAS Models:}
Table~\ref{table:continual-learning-videocaptioning-results} also shows that our ENAS models (MSR-VTT, MSVD) perform equal/better than non-architecture search based models.\footnote{Note that ENAS random search performance on MSR-VTT test set is C:43.3, B:37.0, R:58.7, M:27.3, AVG: 41.6; and on MSVD test set is C:83.7, B:47.4, R:71.1, M:33.6, AVG: 59.0, suggesting that these are lower than the learned optimal cell structures' performances shown in Table~\ref{table:continual-learning-videocaptioning-results}.}

\noindent\textbf{CAS Models:}
Next, we apply our continual architecture search (CAS) approach on MSR-VTT and MSVD, where we sequentially allow the model to learn MSR-VTT first and then MSVD, and the results are as shown in Table~\ref{table:continual-learning-videocaptioning-results}. We observe that even though we learn the models sequentially, we are able to maintain performance on the previously-learned MSR-VTT task in step-2, while also achieving greater-or-equal performance on the current task of MSVD in comparison with the general ENAS approach.\footnote{MSR-VTT performance in step-1 and step-2 are stat. equal on CIDEr and ROUGE-L metrics.} 

\noindent\textbf{Human Evaluation:}
We also performed human comparison of our CAS step-1 vs. step-2 via Amazon MTurk (100 anonymized test samples, Likert 1-5 scale). This gave an overall score of 3.62 for CAS step-1 model vs. 3.55 for CAS step-2, which are very close (statistically insignificant with $p=0.32$), again showing that CAS step-2 is able to maintain performance w.r.t. CAS step-1.

\subsection{Multi-Task Cell Learning on GLUE}
\label{subsec:multi-task-cell-glue}

In these experiments, we first find the best ENAS cell structures for the individual QNLI and WNLI tasks, and use these for training the RTE task. Next, we find a joint cell structure by training ENAS via joint rewards from both QNLI and WNLI datasets. Later, we use this single `multi-task' cell to train the RTE task, and the results are as shown in Table~\ref{table:multitask-cell-glue-results} (GLUE test results). We also include the LSTM cell and RTE-ENAS cell results for fair comparison. It is clear that the multi-task cell performs better than the single-task cells.\footnote{Our multi-task cell and RTE cell performance are statistically equal (61.4 vs. 60.3) and statistically better than the rest of the cells in Table~\ref{table:multitask-cell-glue-results}, based on the validation set. Note that the multi-task cell does not necessarily need to be better than the RTE cell, because the latter cell will be over-optimized for its own data, while the former is a more generalized cell learned from two other datasets.} This shows that a cell learned on multiple tasks is more generalizable to other tasks. 
\subsection{Multi-Task Cell on Video Captioning}
In these experiments, we first find the best ENAS cell structures for the individual MSR-VTT and MSVD tasks, and use these cell structures for training the DiDeMo task. Next, we find a single cell structure by training ENAS on both MSR-VTT and MSVD datasets jointly. Later, we use this single cell (we call it multi-task cell) to train the DiDeMo task, and the results are as shown in Table~\ref{table:multitask-cell-didemo-results}. It is clear that the multi-task cell performs better than other cell structures, where the multi-task cell performance is comparable w.r.t. the DiDeMo-ENAS cell and better than the other single-task and LSTM cell structures. This shows that a cell learned on multiple tasks is more generalizable to other tasks.

\noindent\textbf{Human Evaluation:} 
We performed a similar human study as Sec.~\ref{subsec:continual-learning-on-video-captioning}, and got Likert scores of 2.94 for multi-task cell vs. 2.81 for LSTM cell, which suggests that the multi-task cell is more generalizable than the standard LSTM cell.

\begin{table}[t]
\small
\begin{center}
\begin{tabular}{|l|c|c|c|}
\hline
Cell Structure &  Performance on RTE \\
\hline
LSTM cell & 52.3 \\
QNLI cell & 52.4 \\ 
WNLI cell & 52.2  \\ 
RTE cell & 52.9 \\
Multi-Task cell & 53.9 \\ 
\hline
\end{tabular}
\end{center}
\vspace{-7pt}
\caption{Comparison of MAS cell on RTE task.
\label{table:multitask-cell-glue-results}\vspace{-5pt}}
\end{table}

\begin{table}[t]
\small
\begin{center}
\begin{tabular}{|l|cccc|}
\hline
\multirow{2}{*}{Cell Structure} &  \multicolumn{4}{c|}{Performance on DiDeMo} \\
\cline{2-5}
& M & C & B & R \\
\hline
LSTM cell & 12.7 & 26.7 & 7.6 & 30.6\\
MSR-VTT cell & 12.9 & 25.7 & 7.4 & 30.3 \\ 
MSVD cell & 12.1 & 25.2 & 7.9 & 30.6 \\ 
DiDeMO cell & 13.1 & 27.1 & 7.9 & 30.9 \\
Multi-Task cell & 13.4 & 27.5 & 8.1 & 30.8 \\ 
\hline
\end{tabular}
\end{center}    
\vspace{-7pt}
\caption{Comparison of MAS cell on DiDeMO task.
\label{table:multitask-cell-didemo-results}}
\vspace{-10pt}
\end{table}

\subsection{Analysis}
\label{subsec:analysis}

\begin{figure}[t]
\centering
\begin{subfigure}{.38\linewidth}
  \centering
  \includegraphics[height=2.5cm]{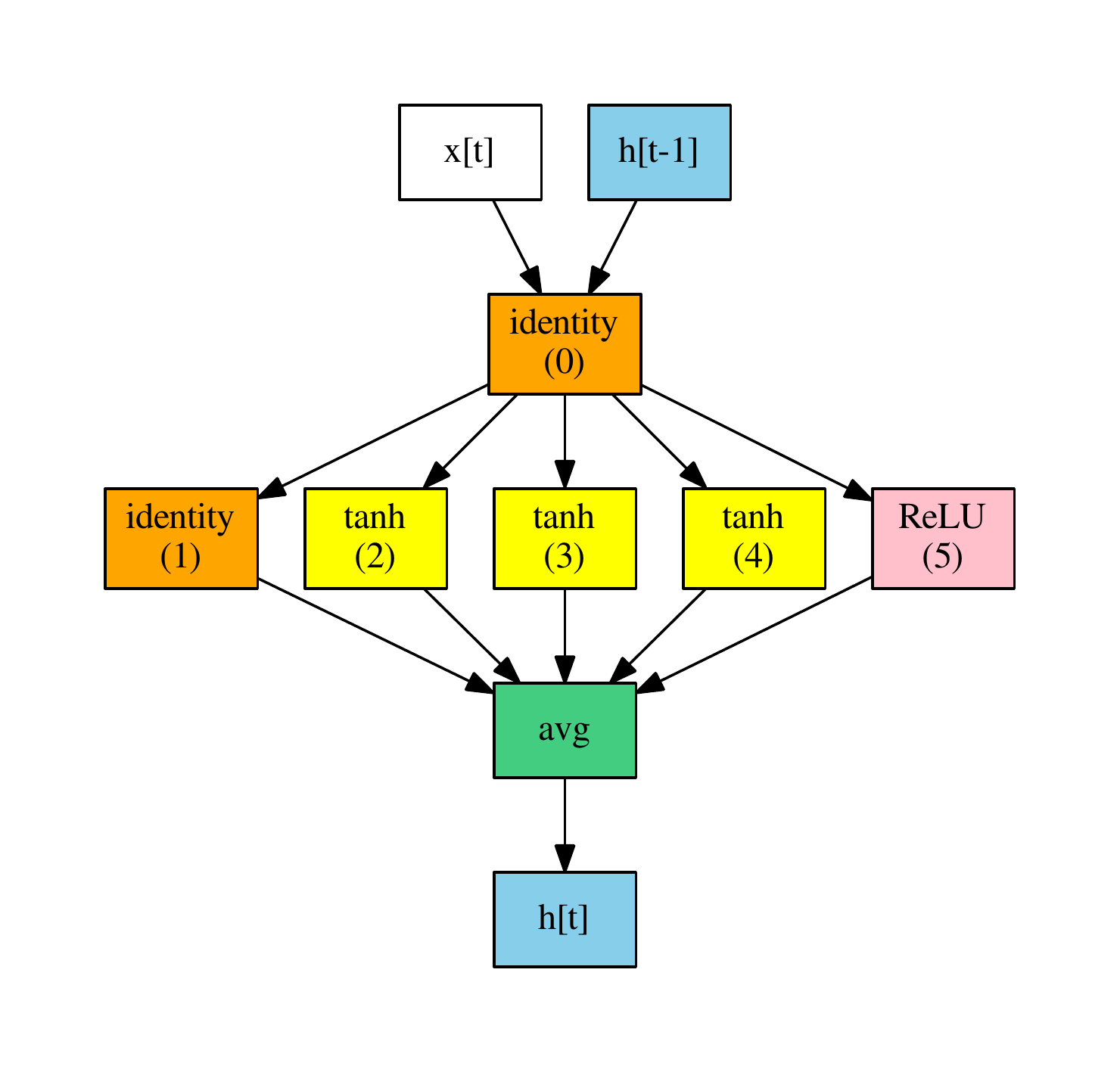}
  \caption{Step-1}
\end{subfigure}
\begin{subfigure}{.28\linewidth}
  \centering
  \includegraphics[height=2.5cm]{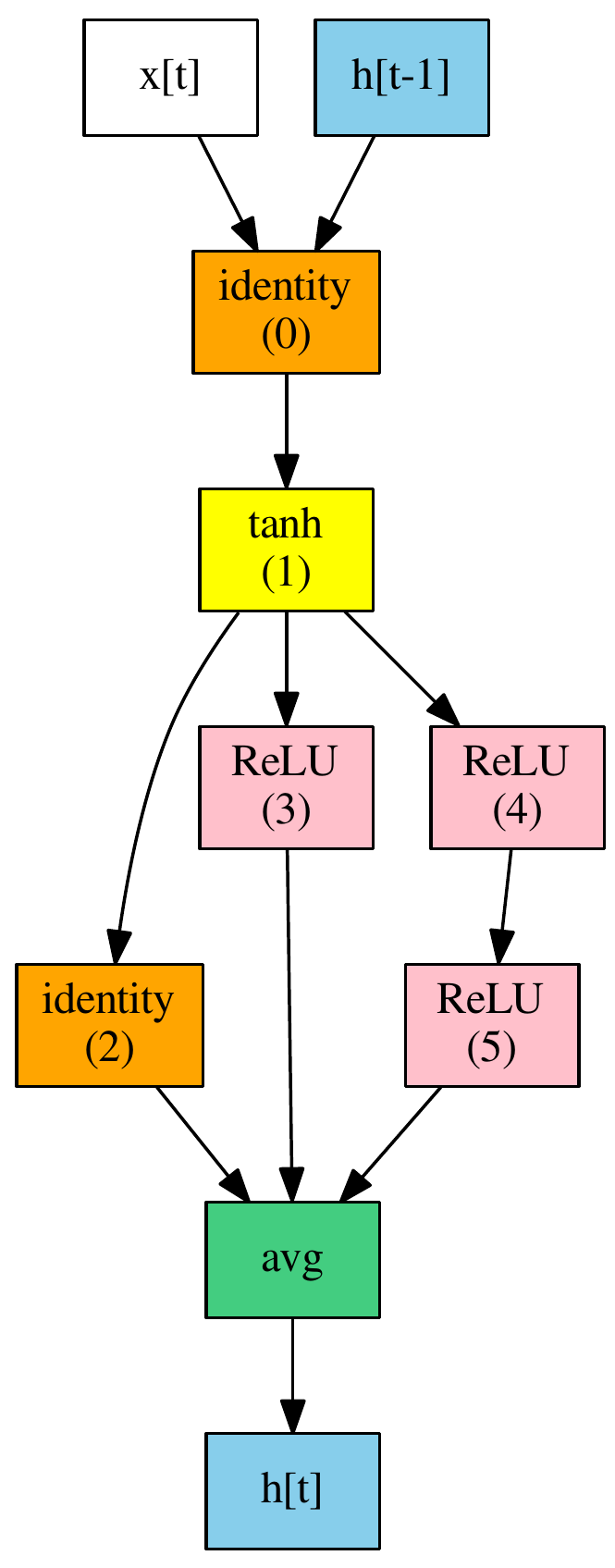}
  \caption{Step-2}
\end{subfigure}
\begin{subfigure}{.28\linewidth}
  \centering
  \includegraphics[height=2.5cm]{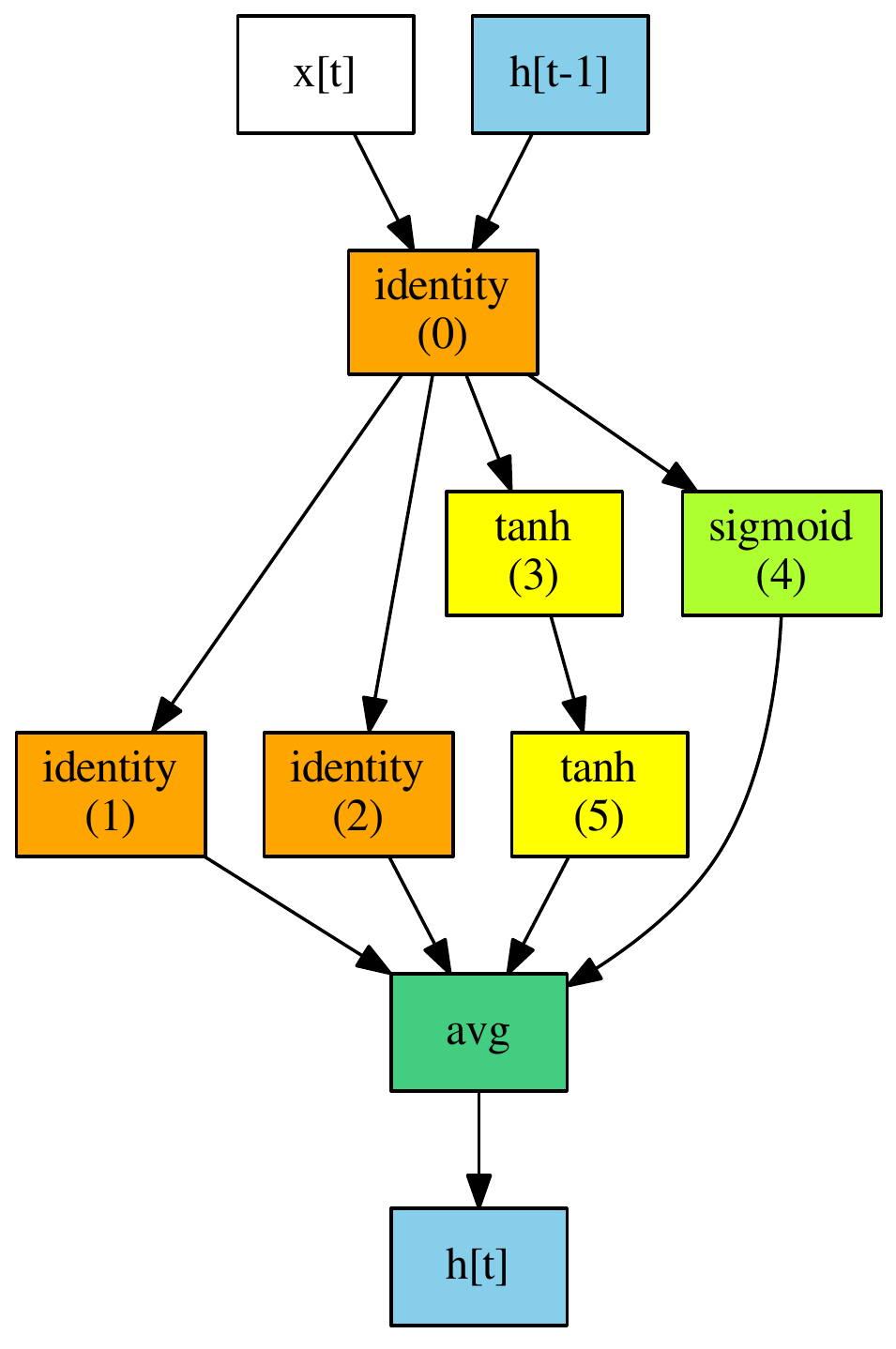}
  \caption{Step-3}
\end{subfigure}
\vspace{-7pt}
\caption{Learned cell structures for step-1, step-2, and step-3 of continual architecture search for GLUE tasks.
\label{fig:cas-cell-strctures}}
\vspace{-14pt}
\end{figure}

\paragraph{Evolved Cell Structure with CAS}
Fig.~\ref{fig:cas-cell-strctures} presents the cell structure in each step for the CAS approach, where we sequentially train QNLI, RTE, and WNLI tasks. Overall, we observe that the cell structures in CAS preserve the properties of certain edges while creating new edges for new capabilities. We notice that the cell structure in step-1 and step-2 share some common edges and activation functions (e.g., inputs to node 0) along with some new edge connections in step-2 (e.g., node 1 to node 3). Further, we observe that the step-3 cell uses some common edges w.r.t. the step-2 cell, but uses different activation functions, e.g., edge between node 0 and node 1 is the same, but the activation function is different. This shows that those edges are learning weights which are stable w.r.t. change in the activation functions.

\begin{figure}[t]

\begin{subfigure}{.48\linewidth}
  \centering
  \includegraphics[height=2.5cm]{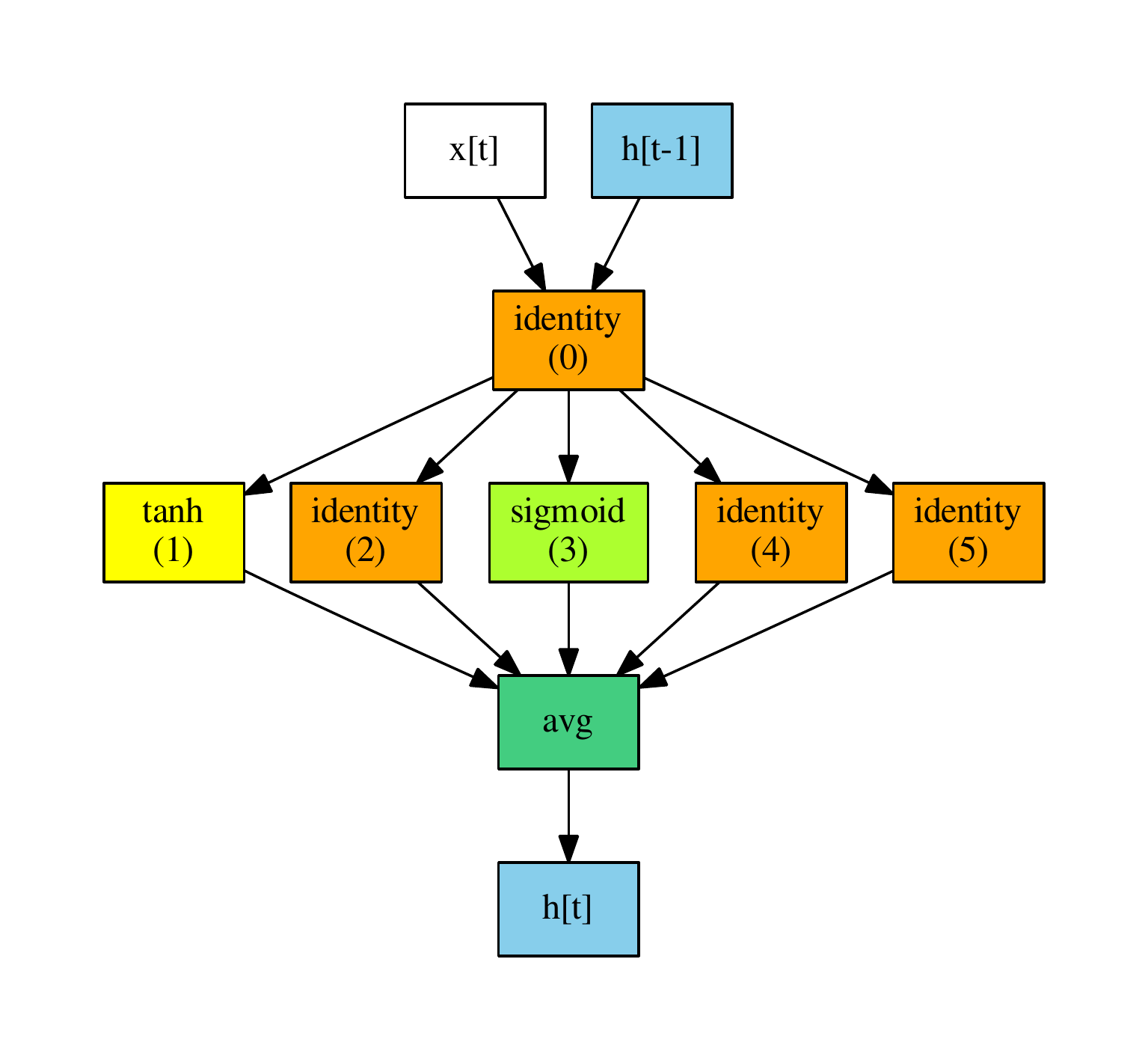}
  \caption{MAS cell}
\end{subfigure}
\begin{subfigure}{.48\linewidth}
  \centering
  \includegraphics[height=2.5cm]{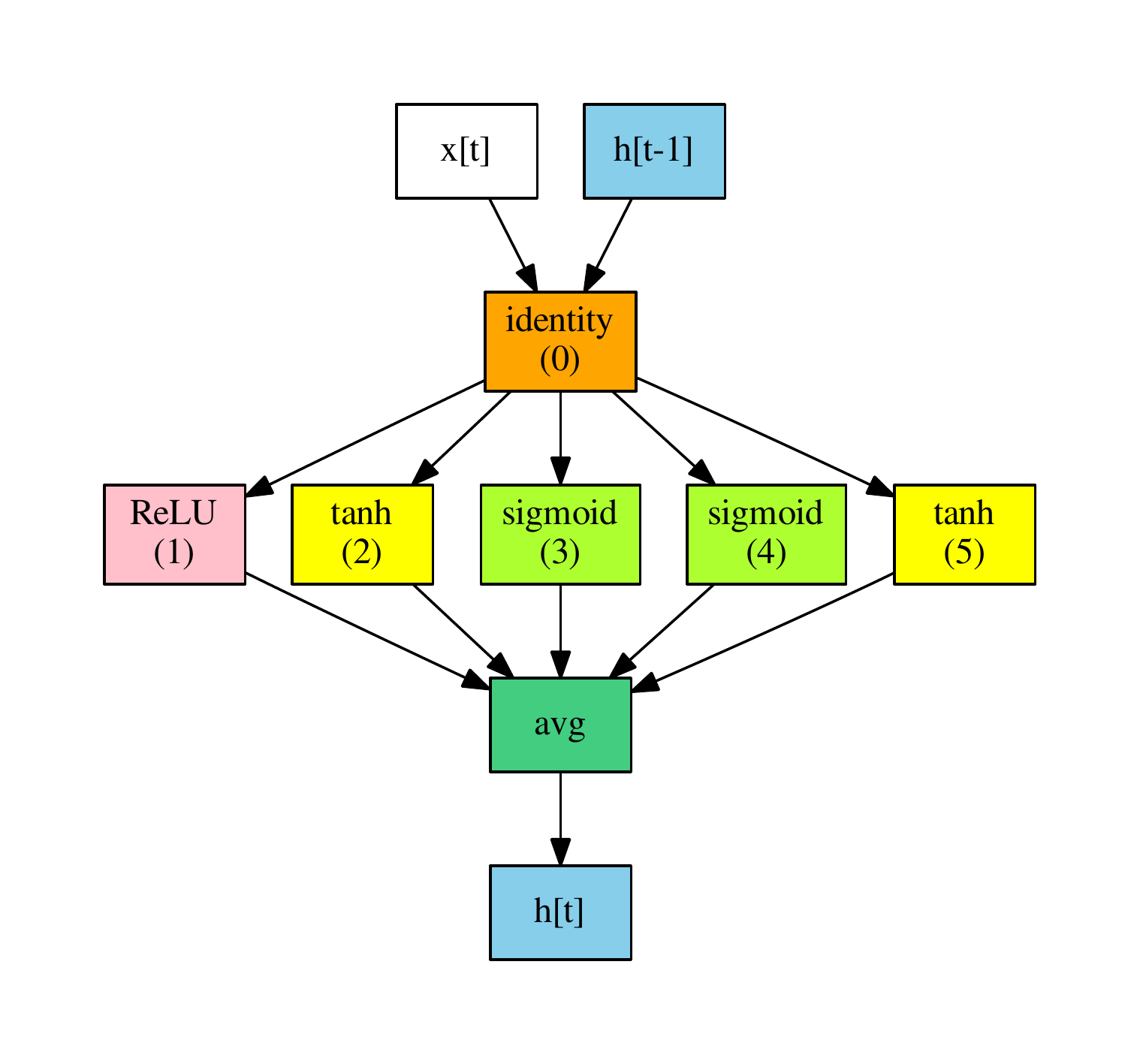}
  \caption{RTE cell}
\end{subfigure}
\vspace{-7pt}
    \caption{Learned multi-task \& RTE cell structures.}
    \label{fig:multitask-cell-strcture}
    \vspace{-14pt}
\end{figure}

\paragraph{Multi-Task Cell Structure}
Fig.~\ref{fig:multitask-cell-strcture} presents our multi-task MAS cell structure (with joint rewards from QNLI and WNLI), versus the RTE-ENAS cell structure. We observe that the MAS cell is relatively less complex, i.e., uses several identity functions and very few activation functions in its structure vs. the RTE cell. This shows that the individual-task-optimized cell structures are complex and over-specialized to that task, whereas our multi-task cell structures are simpler for generalizability to new unseen tasks.

\section{Conclusion}
We first presented an architecture search approach for text classification and video caption generation tasks. Next, we introduced a novel paradigm of transfer learning by combining architecture search with continual learning to avoid catastrophic forgetting. We also explore multi-task cell learning for generalizability. 

\section*{Acknowledgments}
We thank the reviewers for their helpful comments. This work was supported by DARPA (YFA17-D17AP00022), and faculty awards from Google, Facebook, and Salesforce.  The views contained in this article are those of the authors and not of the funding agency.

\bibliography{main}

\begin{thebibliography}{59}
\expandafter\ifx\csname natexlab\endcsname\relax\def\natexlab#1{#1}\fi

\bibitem[{Andreas et~al.(2016)Andreas, Rohrbach, Darrell, and
  Klein}]{andreas2016learning}
Jacob Andreas, Marcus Rohrbach, Trevor Darrell, and Dan Klein. 2016.
\newblock Learning to compose neural networks for question answering.
\newblock In \emph{NAACL}.

\bibitem[{Augenstein et~al.(2018)Augenstein, Ruder, and
  S{\o}gaard}]{augenstein2018multi}
Isabelle Augenstein, Sebastian Ruder, and Anders S{\o}gaard. 2018.
\newblock Multi-task learning of pairwise sequence classification tasks over
  disparate label spaces.
\newblock In \emph{NAACL}.

\bibitem[{Bahdanau et~al.(2015)Bahdanau, Cho, and Bengio}]{bahdanau2014neural}
Dzmitry Bahdanau, Kyunghyun Cho, and Yoshua Bengio. 2015.
\newblock Neural machine translation by jointly learning to align and
  translate.
\newblock In \emph{ICLR}.

\bibitem[{Baker et~al.(2017)Baker, Gupta, Naik, and
  Raskar}]{baker2016designing}
Bowen Baker, Otkrist Gupta, Nikhil Naik, and Ramesh Raskar. 2017.
\newblock Designing neural network architectures using reinforcement learning.
\newblock In \emph{ICLR}.

\bibitem[{Biermann(1978)}]{biermann1978inference}
Alan~W Biermann. 1978.
\newblock The inference of regular lisp programs from examples.
\newblock \emph{IEEE transactions on Systems, Man, and Cybernetics},
  8(8):585--600.

\bibitem[{Bousmalis et~al.(2016)Bousmalis, Trigeorgis, Silberman, Krishnan, and
  Erhan}]{bousmalis2016domain}
Konstantinos Bousmalis, George Trigeorgis, Nathan Silberman, Dilip Krishnan,
  and Dumitru Erhan. 2016.
\newblock Domain separation networks.
\newblock In \emph{NIPS}, pages 343--351.

\bibitem[{Cai et~al.(2018)Cai, Chen, Zhang, Yu, and Wang}]{cai2018efficient}
Han Cai, Tianyao Chen, Weinan Zhang, Yong Yu, and Jun Wang. 2018.
\newblock Efficient architecture search by network transformation.
\newblock In \emph{AAAI}.

\bibitem[{Caruana(1998)}]{caruana1998multitask}
Rich Caruana. 1998.
\newblock Multitask learning.
\newblock In \emph{Learning to learn}, pages 95--133. Springer.

\bibitem[{Chen and Dolan(2011)}]{chen2011collecting}
David~L Chen and William~B Dolan. 2011.
\newblock Collecting highly parallel data for paraphrase evaluation.
\newblock In \emph{ACL}.

\bibitem[{Chen et~al.(2015)Chen, Fang, Lin, Vedantam, Gupta, Doll{\'a}r, and
  Zitnick}]{chen2015microsoft}
Xinlei Chen, Hao Fang, Tsung-Yi Lin, Ramakrishna Vedantam, Saurabh Gupta, Piotr
  Doll{\'a}r, and C~Lawrence Zitnick. 2015.
\newblock Microsoft {COCO} captions: Data collection and evaluation server.
\newblock \emph{arXiv preprint arXiv:1504.00325}.

\bibitem[{Chen and Liu(2016)}]{chen2016lifelong}
Zhiyuan Chen and Bing Liu. 2016.
\newblock Lifelong machine learning.
\newblock \emph{Synthesis Lectures on Artificial Intelligence and Machine
  Learning}, 10(3):1--145.

\bibitem[{Collobert and Weston(2008)}]{collobert2008unified}
Ronan Collobert and Jason Weston. 2008.
\newblock A unified architecture for natural language processing: Deep neural
  networks with multitask learning.
\newblock In \emph{Proceedings of the 25th international conference on Machine
  learning}, pages 160--167. ACM.

\bibitem[{Conneau et~al.(2017)Conneau, Kiela, Schwenk, Barrault, and
  Bordes}]{conneau2017supervised}
Alexis Conneau, Douwe Kiela, Holger Schwenk, Loic Barrault, and Antoine Bordes.
  2017.
\newblock Supervised learning of universal sentence representations from
  natural language inference data.
\newblock In \emph{EMNLP}.

\bibitem[{Denkowski and Lavie(2014)}]{banerjee2005meteor}
Michael Denkowski and Alon Lavie. 2014.
\newblock Meteor universal: Language specific translation evaluation for any
  target language.
\newblock In \emph{EACL}.

\bibitem[{Donahue et~al.(2014)Donahue, Jia, Vinyals, Hoffman, Zhang, Tzeng, and
  Darrell}]{donahue2014decaf}
Jeff Donahue, Yangqing Jia, Oriol Vinyals, Judy Hoffman, Ning Zhang, Eric
  Tzeng, and Trevor Darrell. 2014.
\newblock Decaf: A deep convolutional activation feature for generic visual
  recognition.
\newblock In \emph{ICML}, pages 647--655.

\bibitem[{Efron and Tibshirani(1994)}]{efron1994introduction}
Bradley Efron and Robert~J Tibshirani. 1994.
\newblock \emph{An introduction to the bootstrap}.
\newblock CRC press.

\bibitem[{Elsken et~al.(2019)Elsken, Metzen, and Hutter}]{elsken2018efficient}
Thomas Elsken, Jan~Hendrik Metzen, and Frank Hutter. 2019.
\newblock Efficient multi-objective neural architecture search via lamarckian
  evolution.
\newblock In \emph{ICLR}.

\bibitem[{Girshick(2015)}]{girshick2015fast}
Ross Girshick. 2015.
\newblock Fast r-cnn.
\newblock In \emph{Proceedings of the IEEE international conference on computer
  vision}, pages 1440--1448.

\bibitem[{Guo et~al.(2018)Guo, Pasunuru, and Bansal}]{guo2018soft}
Han Guo, Ramakanth Pasunuru, and Mohit Bansal. 2018.
\newblock Soft layer-specific multi-task summarization with entailment and
  question generation.
\newblock In \emph{ACL}.

\bibitem[{He et~al.(2016)He, Zhang, Ren, and Sun}]{he2016deep}
Kaiming He, Xiangyu Zhang, Shaoqing Ren, and Jian Sun. 2016.
\newblock Deep residual learning for image recognition.
\newblock In \emph{CVPR}, pages 770--778.

\bibitem[{Hendricks et~al.(2017)Hendricks, Wang, Shechtman, Sivic, Darrell, and
  Russell}]{hendricks2017localizing}
Lisa~Anne Hendricks, Oliver Wang, Eli Shechtman, Josef Sivic, Trevor Darrell,
  and Bryan Russell. 2017.
\newblock Localizing moments in video with natural language.
\newblock In \emph{ICCV}, pages 5803--5812.

\bibitem[{Hinton et~al.(2015)Hinton, Vinyals, and Dean}]{hinton2015distilling}
Geoffrey Hinton, Oriol Vinyals, and Jeff Dean. 2015.
\newblock Distilling the knowledge in a neural network.
\newblock \emph{arXiv preprint arXiv:1503.02531}.

\bibitem[{Jung et~al.(2016)Jung, Ju, Jung, and Kim}]{jung2016less}
Heechul Jung, Jeongwoo Ju, Minju Jung, and Junmo Kim. 2016.
\newblock Less-forgetting learning in deep neural networks.
\newblock \emph{arXiv preprint arXiv:1607.00122}.

\bibitem[{Kingma and Ba(2015)}]{kingma2014adam}
Diederik Kingma and Jimmy Ba. 2015.
\newblock Adam: A method for stochastic optimization.
\newblock In \emph{ICLR}.

\bibitem[{Kirkpatrick et~al.(2017)Kirkpatrick, Pascanu, Rabinowitz, Veness,
  Desjardins, Rusu, Milan, Quan, Ramalho, Grabska-Barwinska
  et~al.}]{kirkpatrick2017overcoming}
James Kirkpatrick, Razvan Pascanu, Neil Rabinowitz, Joel Veness, Guillaume
  Desjardins, Andrei~A Rusu, Kieran Milan, John Quan, Tiago Ramalho, Agnieszka
  Grabska-Barwinska, et~al. 2017.
\newblock Overcoming catastrophic forgetting in neural networks.
\newblock \emph{Proceedings of the National Academy of Sciences},
  114(13):3521--3526.

\bibitem[{Lake et~al.(2015)Lake, Salakhutdinov, and Tenenbaum}]{lake2015human}
Brenden~M Lake, Ruslan Salakhutdinov, and Joshua~B Tenenbaum. 2015.
\newblock Human-level concept learning through probabilistic program induction.
\newblock \emph{Science}, 350(6266):1332--1338.

\bibitem[{Li and Hoiem(2017)}]{li2017learning}
Zhizhong Li and Derek Hoiem. 2017.
\newblock Learning without forgetting.
\newblock \emph{IEEE Transactions on Pattern Analysis and Machine
  Intelligence}.

\bibitem[{Liang et~al.(2010)Liang, Jordan, and Klein}]{liang2010learning}
Percy Liang, Michael~I Jordan, and Dan Klein. 2010.
\newblock Learning programs: A hierarchical bayesian approach.
\newblock In \emph{ICML}, pages 639--646.

\bibitem[{Lin(2004)}]{lin2004rouge}
Chin-Yew Lin. 2004.
\newblock {ROUGE}: A package for automatic evaluation of summaries.
\newblock In \emph{Text Summarization Branches Out: Proceedings of the ACL-04
  workshop}, volume~8.

\bibitem[{Liu et~al.(2017)Liu, Zoph, Shlens, Hua, Li, Fei-Fei, Yuille, Huang,
  and Murphy}]{liu2017progressive}
Chenxi Liu, Barret Zoph, Jonathon Shlens, Wei Hua, Li-Jia Li, Li~Fei-Fei, Alan
  Yuille, Jonathan Huang, and Kevin Murphy. 2017.
\newblock Progressive neural architecture search.
\newblock \emph{arXiv preprint arXiv:1712.00559}.

\bibitem[{Liu et~al.(2018)Liu, Simonyan, Vinyals, Fernando, and
  Kavukcuoglu}]{liu2017hierarchical}
Hanxiao Liu, Karen Simonyan, Oriol Vinyals, Chrisantha Fernando, and Koray
  Kavukcuoglu. 2018.
\newblock Hierarchical representations for efficient architecture search.
\newblock In \emph{CVPR}.

\bibitem[{Luong et~al.(2015)Luong, Le, Sutskever, Vinyals, and
  Kaiser}]{luong2015multi}
Minh-Thang Luong, Quoc~V Le, Ilya Sutskever, Oriol Vinyals, and Lukasz Kaiser.
  2015.
\newblock Multi-task sequence to sequence learning.
\newblock \emph{arXiv preprint arXiv:1511.06114}.

\bibitem[{Neelakantan et~al.(2015)Neelakantan, Le, and
  Sutskever}]{neelakantan2015neural}
Arvind Neelakantan, Quoc~V Le, and Ilya Sutskever. 2015.
\newblock Neural programmer: Inducing latent programs with gradient descent.
\newblock In \emph{ICLR}.

\bibitem[{Negrinho and Gordon(2017)}]{negrinho2017deeparchitect}
Renato Negrinho and Geoff Gordon. 2017.
\newblock Deeparchitect: Automatically designing and training deep
  architectures.
\newblock In \emph{CVPR}.

\bibitem[{Nguyen et~al.(2017)Nguyen, Li, Bui, and
  Turner}]{nguyen2017variational}
Cuong~V Nguyen, Yingzhen Li, Thang~D Bui, and Richard~E Turner. 2017.
\newblock Variational continual learning.
\newblock \emph{arXiv preprint arXiv:1710.10628}.

\bibitem[{Noreen(1989)}]{noreen1989computer}
Eric~W Noreen. 1989.
\newblock \emph{Computer-intensive methods for testing hypotheses}.
\newblock Wiley New York.

\bibitem[{Oh et~al.(2017)Oh, Singh, Lee, and Kohli}]{oh2017zero}
Junhyuk Oh, Satinder Singh, Honglak Lee, and Pushmeet Kohli. 2017.
\newblock Zero-shot task generalization with multi-task deep reinforcement
  learning.
\newblock \emph{arXiv preprint arXiv:1706.05064}.

\bibitem[{Papineni et~al.(2002)Papineni, Roukos, Ward, and
  Zhu}]{papineni2002bleu}
Kishore Papineni, Salim Roukos, Todd Ward, and Wei-Jing Zhu. 2002.
\newblock {BLEU}: a method for automatic evaluation of machine translation.
\newblock In \emph{ACL}, pages 311--318.

\bibitem[{Pasunuru and Bansal(2017{\natexlab{a}})}]{pasunuru2017multitask}
Ramakanth Pasunuru and Mohit Bansal. 2017{\natexlab{a}}.
\newblock Multi-task video captioning with video and entailment generation.
\newblock In \emph{ACL}.

\bibitem[{Pasunuru and Bansal(2017{\natexlab{b}})}]{pasunuru2017reinforced}
Ramakanth Pasunuru and Mohit Bansal. 2017{\natexlab{b}}.
\newblock Reinforced video captioning with entailment rewards.
\newblock In \emph{EMNLP}.

\bibitem[{Peters et~al.(2018)Peters, Neumann, Iyyer, Gardner, Clark, Lee, and
  Zettlemoyer}]{peters2018deep}
Mat thew~E Peters, Mark Neumann, Mohit Iyyer, Matt Gardner, Christopher Clark,
  Kenton Lee, and Luke Zettlemoyer. 2018.
\newblock Deep contextualized word representations.
\newblock In \emph{NAACL}.

\bibitem[{Pham et~al.(2018)Pham, Guan, Zoph, Le, and Dean}]{pham2018efficient}
Hieu Pham, Melody~Y Guan, Barret Zoph, Quoc~V Le, and Jeff Dean. 2018.
\newblock Efficient neural architecture search via parameter sharing.
\newblock \emph{arXiv preprint arXiv:1802.03268}.

\bibitem[{Rajpurkar et~al.(2016)Rajpurkar, Zhang, Lopyrev, and
  Liang}]{rajpurkar2016squad}
Pranav Rajpurkar, Jian Zhang, Konstantin Lopyrev, and Percy Liang. 2016.
\newblock Squad: 100,000+ questions for machine comprehension of text.
\newblock In \emph{EMNLP}.

\bibitem[{Razavian et~al.(2014)Razavian, Azizpour, Sullivan, and
  Carlsson}]{razavian2014cnn}
Ali~Sharif Razavian, Hossein Azizpour, Josephine Sullivan, and Stefan Carlsson.
  2014.
\newblock Cnn features off-the-shelf: an astounding baseline for recognition.
\newblock In \emph{Computer Vision and Pattern Recognition Workshops (CVPRW),
  2014 IEEE Conference on}, pages 512--519. IEEE.

\bibitem[{Ruder et~al.(2017)Ruder, Bingel, Augenstein, and
  S{\o}gaard}]{ruder2017sluice}
Sebastian Ruder, Joachim Bingel, Isabelle Augenstein, and Anders S{\o}gaard.
  2017.
\newblock Sluice networks: Learning what to share between loosely related
  tasks.
\newblock \emph{arXiv preprint arXiv:1705.08142}.

\bibitem[{Ruder and Plank(2017)}]{ruder2017learning}
Sebastian Ruder and Barbara Plank. 2017.
\newblock Learning to select data for transfer learning with bayesian
  optimization.
\newblock \emph{arXiv preprint arXiv:1707.05246}.

\bibitem[{Rusu et~al.(2016)Rusu, Rabinowitz, Desjardins, Soyer, Kirkpatrick,
  Kavukcuoglu, Pascanu, and Hadsell}]{rusu2016progressive}
Andrei~A Rusu, Neil~C Rabinowitz, Guillaume Desjardins, Hubert Soyer, James
  Kirkpatrick, Koray Kavukcuoglu, Razvan Pascanu, and Raia Hadsell. 2016.
\newblock Progressive neural networks.
\newblock \emph{arXiv preprint arXiv:1606.04671}.

\bibitem[{Scardapane et~al.(2017)Scardapane, Comminiello, Hussain, and
  Uncini}]{scardapane2017group}
Simone Scardapane, Danilo Comminiello, Amir Hussain, and Aurelio Uncini. 2017.
\newblock Group sparse regularization for deep neural networks.
\newblock \emph{Neurocomputing}, 241:81--89.

\bibitem[{So et~al.(2019)So, Liang, and Le}]{so2019evolved}
David~R So, Chen Liang, and Quoc~V Le. 2019.
\newblock The evolved transformer.
\newblock \emph{arXiv preprint arXiv:1901.11117}.

\bibitem[{Summers(1986)}]{summers1986methodology}
Phillip~D Summers. 1986.
\newblock A methodology for lisp program construction from examples.
\newblock In \emph{Readings in artificial intelligence and software
  engineering}, pages 309--316. Elsevier.

\bibitem[{Vedantam et~al.(2015)Vedantam, Lawrence~Zitnick, and
  Parikh}]{vedantam2015cider}
Ramakrishna Vedantam, C~Lawrence~Zitnick, and Devi Parikh. 2015.
\newblock {CIDE}r: Consensus-based image description evaluation.
\newblock In \emph{CVPR}, pages 4566--4575.

\bibitem[{Wang et~al.(2018)Wang, Singh, Michael, Hill, Levy, and
  Bowman}]{wang2018glue}
Alex Wang, Amapreet Singh, Julian Michael, Felix Hill, Omer Levy, and Samuel~R
  Bowman. 2018.
\newblock {GLUE}: A multi-task benchmark and analysis platform for natural
  language understanding.
\newblock \emph{arXiv preprint arXiv:1804.07461}.

\bibitem[{Xie et~al.(2017)Xie, Girshick, Doll{\'a}r, Tu, and
  He}]{xie2017aggregated}
Saining Xie, Ross Girshick, Piotr Doll{\'a}r, Zhuowen Tu, and Kaiming He. 2017.
\newblock Aggregated residual transformations for deep neural networks.
\newblock In \emph{CVPR}, pages 5987--5995. IEEE.

\bibitem[{Xu et~al.(2016)Xu, Mei, Yao, and Rui}]{xu2016msr}
Jun Xu, Tao Mei, Ting Yao, and Yong Rui. 2016.
\newblock {MSR-VTT}: A large video description dataset for bridging video and
  language.
\newblock In \emph{CVPR}, pages 5288--5296. IEEE.

\bibitem[{Yoon et~al.(2018)Yoon, Yang, Lee, and Hwang}]{yoon2018lifelong}
Jaehong Yoon, Eunho Yang, Jeongtae Lee, and Sung~Ju Hwang. 2018.
\newblock Lifelong learning with dynamically expandable networks.
\newblock \emph{arXiv preprint arXiv:1708.01547}.

\bibitem[{Yosinski et~al.(2014)Yosinski, Clune, Bengio, and
  Lipson}]{yosinski2014transferable}
Jason Yosinski, Jeff Clune, Yoshua Bengio, and Hod Lipson. 2014.
\newblock How transferable are features in deep neural networks?
\newblock In \emph{NIPS}, pages 3320--3328.

\bibitem[{Zenke et~al.(2017)Zenke, Poole, and Ganguli}]{zenke2017continual}
Friedemann Zenke, Ben Poole, and Surya Ganguli. 2017.
\newblock Continual learning through synaptic intelligence.
\newblock In \emph{ICML}, pages 3987--3995.

\bibitem[{Zoph and Le(2017)}]{zoph2016neural}
Barret Zoph and Quoc~V Le. 2017.
\newblock Neural architecture search with reinforcement learning.
\newblock In \emph{ICLR}.

\bibitem[{Zoph et~al.(2018)Zoph, Vasudevan, Shlens, and Le}]{zoph2017learning}
Barret Zoph, Vijay Vasudevan, Jonathon Shlens, and Quoc~V Le. 2018.
\newblock Learning transferable architectures for scalable image recognition.
\newblock In \emph{CVPR}.

\end{thebibliography}
\bibliographystyle{acl_natbib}

\appendix

\section*{Appendix}

\section{Training Details}
We use Adam optimizer~\cite{kingma2014adam} and a mini-batch size of 64. We set the dropout to 0.5. In all of our architecture search models, we use 6 nodes. For the controller's optimization, we again use Adam optimizer with a learning rate of 0.00035. 

For GLUE tasks, we use 256 dimensions for the hidden states of the RNNs, and for word embeddings we use ELMo representations~\cite{peters2018deep}, where we down project the 1024 dimensions ELMo embeddings to 256. We use a learning rate of 0.001, and both encoder RNNs are unrolled to 50 steps. For CAS conditions, we set the coefficients for block-sparsity and orthogonality conditions to 0.001 and 0.001, respectively. 

For video captioning tasks, we use hidden state size of 1024 and word embedding size of 512. For visual features, we use a concatenation of both ResNet-152~\cite{he2016deep} and ResNeXt-101~\cite{xie2017aggregated} image features. We use a learning rate of 0.0001, and we unroll the video encoder and caption decoder to 50 and 20 steps, respectively. For CAS conditions, we set both the coefficients of block-sparsity and orthogonality conditions to 0.0001.

\end{document}